\documentclass[10pt,twocolumn,letterpaper]{article}

\usepackage[pagenumbers]{cvpr} 

\definecolor{cvprblue}{rgb}{0.21,0.49,0.74}
\usepackage[pagebackref,breaklinks,colorlinks,allcolors=cvprblue]{hyperref}

\def\paperID{21} 
\def\confName{CVPR}
\def\confYear{2025}

\title{CameraBench: Benchmarking Visual Reasoning in MLLMs via Photography}

\author{I-Sheng Fang\quad Jun-Cheng Chen\\
Research Center for Information Technology Innovation, Academia Sinica \\
{\tt\small \{ishengfang, pullpull\}@citi.sinica.edu.tw}
}

\begin{document}
\maketitle
\begin{abstract}
Large language models (LLMs) and multimodal large language models (MLLMs) have significantly advanced artificial intelligence. However, visual reasoning, reasoning involving both visual and textual inputs, remains underexplored. Recent advancements, including the reasoning models like OpenAI o1 and Gemini 2.0 Flash Thinking, which incorporate image inputs, have opened this capability. In this ongoing work, we focus specifically on photography-related tasks because a photo is a visual snapshot of the physical world where the underlying physics (i.e., illumination, blur extent, etc.) interplay with the camera parameters. Successfully reasoning from the visual information of a photo to identify these numerical camera settings requires the MLLMs to have a deeper understanding of the underlying physics for precise visual comprehension, representing a challenging and intelligent capability essential for practical applications like photography assistant agents. We aim to evaluate MLLMs on their ability to distinguish visual differences related to numerical camera settings, extending a methodology previously proposed for vision-language models (VLMs). Our preliminary results demonstrate the importance of visual reasoning in photography-related tasks. Moreover, these results show that no single MLLM consistently dominates across all evaluation tasks, demonstrating ongoing challenges and opportunities in developing MLLMs with better visual reasoning.
\end{abstract}
\vspace{-3em}    

\begin{figure}
    \centering
    \includegraphics[width=\columnwidth]{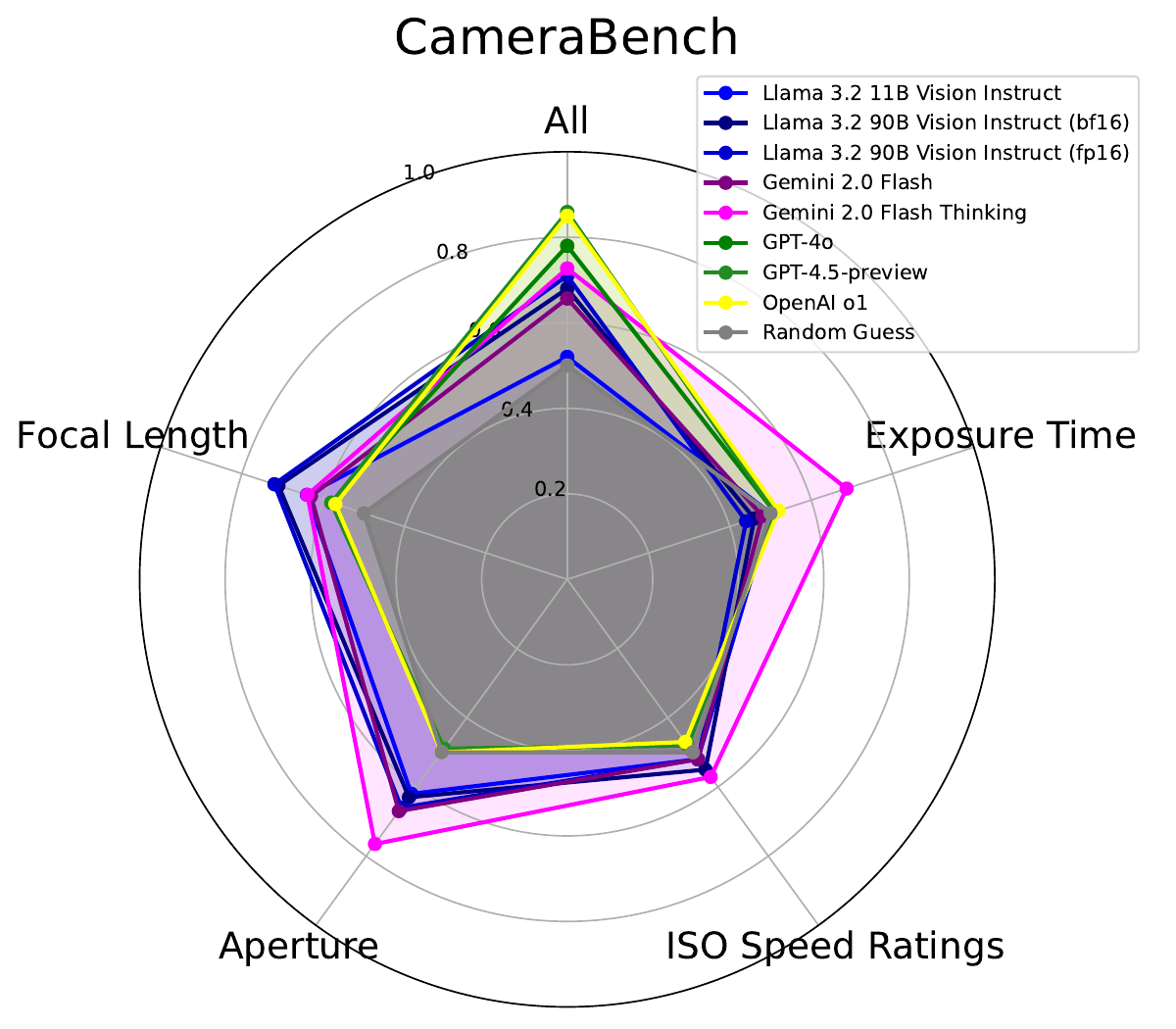}
    \vspace{-1em}
    \caption{ \textbf{Radar chart of CameraBench.}
    We evaluate MLLMs by performing binary-choice questions on a set of numerical camera settings associated with the given images. The questions are conducted under different conditions.
    All: The camera settings in the two options are all different.
    Focal Length: Only the focal length is different between the two options.
    Aperture: Only the aperture is different between the two options.
    ISO: Only the ISO speed rating is different between the two options.
    Exposure Time: Only the exposure time is different between the two options.
    }
    \label{fig:radar-chart}
    \vspace{-1em}
\end{figure}

\section{Introduction}
\label{sec:intro}

Large language models (LLMs)~\cite{brown2020gpt3, achiam2023gpt, touvron2023llama, dubey2024llama} and multimodal large language models (MLLMs)~\cite{hurst2024gpt4o, liu2023llava, team2023gemini, team2024gemini, meta2024llama32} have become dominant forces in artificial intelligence research. Recent advancements, such as OpenAI o1~\cite{jaech2024o1}, Gemini 2.0 Flash Thinking~\cite{team2023gemini}, and DeepSeek R1~\cite{guo2025deepseek}, have further pushed the field forward by incorporating reasoning capabilities, significantly improving the performance in reasoning tasks. Despite these advances, \textit{visual reasoning}, the ability to reason based on both visual and textual inputs, remains underexplored because many reasoning LLMs only use text as the input. 
However, recent models, such as OpenAI o1, and Gemini 2.0 Flash Thinking, now support image inputs while DeepSeek R1 only supports text input, making visual reasoning an emerging research direction.

\begin{figure}[th]
    \centering
    \resizebox{\columnwidth}{!}{
    \begin{tabular}{c}
    \parbox[c]{\columnwidth}{
    \includegraphics[width=\columnwidth]{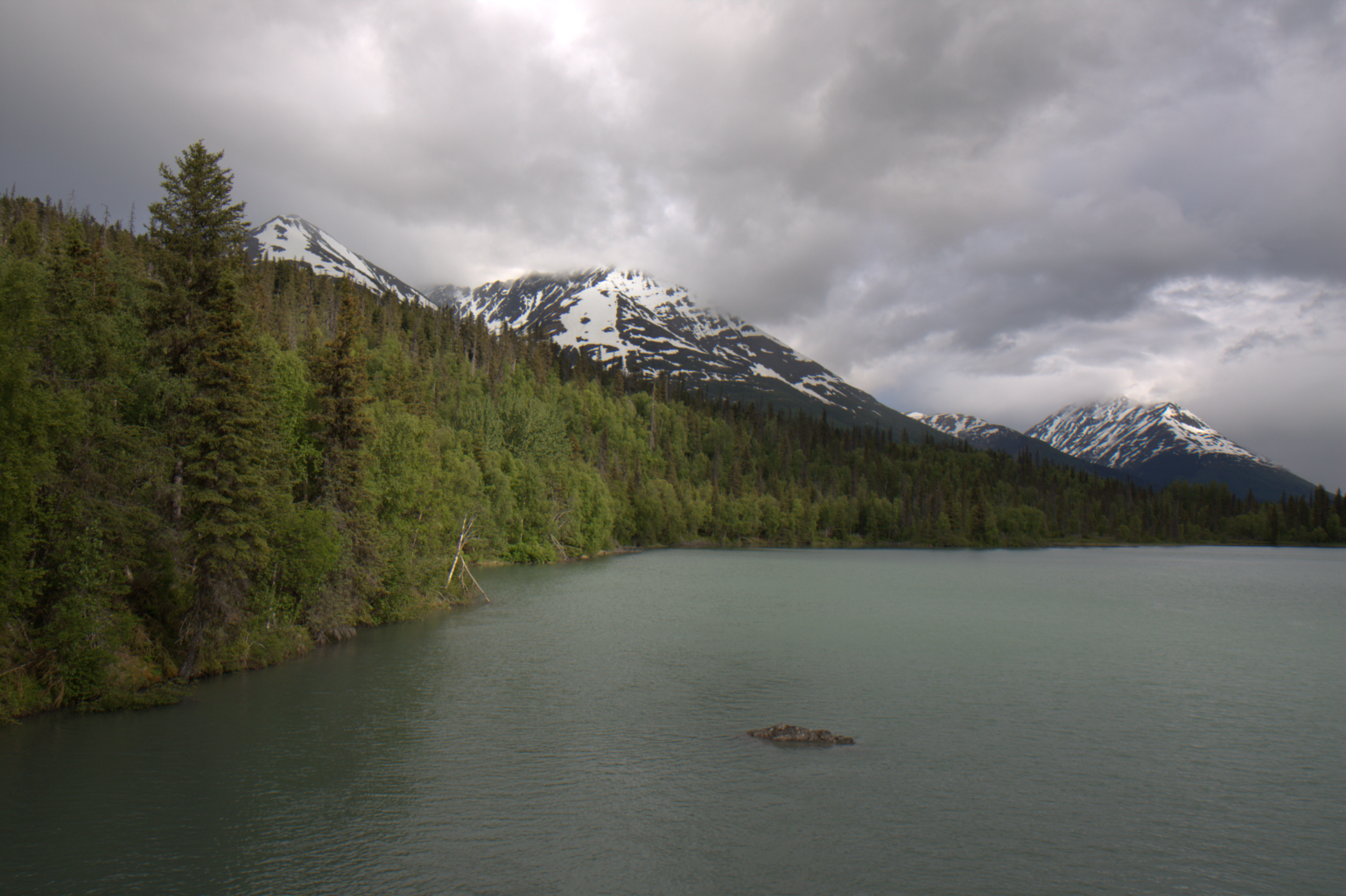}
    }
    \\
    \textbf{Task (1) Prompt} \\
    \hline
    \resizebox{\columnwidth}{!}{
    \begin{tabular}{l}
    Which set of camera settings is most likely to be used in \\
    this image with a full-frame camera? \\
    (A) Focal length: 27.54mm, Aperture: f/14.58, ISO: 524.88, \\
    Exposure Time: 0.00625 second \\
    (B) Focal length: 50.00mm, Aperture: f/4.50, ISO: 200.00, \\
    Exposure Time: 0.0125 second \\
    \\
    Output answer as following format. \\
    \\
    **Answer**: (option) \\
    \end{tabular}
    }
    \\
    \hline
    \textbf{Task (6) Prompt} \\
    \hline
    \resizebox{\columnwidth}{!}{
    \begin{tabular}{l}
    Which set of camera settings is most likely to be used in \\
    this image with a full-frame camera? \\
    (A) Focal length: 27.54mm, Aperture: f/4.50, ISO: 524.88, \\ Exposure Time: 0.00625 second \\
    (B) Focal length: 27.54mm, Aperture: f/14.58, ISO: 524.88, \\
    Exposure Time: 0.00625 second \\
    (C) Focal length: 27.54mm, Aperture: f/14.58, ISO: 200.00, \\
    Exposure Time: 0.00625 second \\
    (D) Focal length: 50.00mm, Aperture: f/14.58, ISO: 524.88, \\
    Exposure Time: 0.00625 second \\
    (E) Focal length: 27.54mm, Aperture: f/14.58, ISO: 524.88, \\
    Exposure Time: 0.0125 second \\
    \\
    Output answer as following format. \\
    \\
    **Answer**: (option) \\
    \hline
    \end{tabular}
    }
    \end{tabular}
    }
    \vspace{-0.5em}
    \caption{\textbf{Example question in CameraBench.} We provide two options of numerical camera settings, one is extracted from the exif of the raw image, another is sampled from other image in CameraSettings25K.}
    \label{fig:sample}
    \vspace{-2em}
\end{figure}

In this ongoing work, we focus on photography-related tasks and introduce Camera Settings Benchmark~(CameraBench) because photography bridges visual, physical, and numerical concepts, providing a practical and ideal approach to evaluating visual reasoning. Specifically, adjusting numerical camera settings directly influences the visual appearance of photographs. For example, a photograph's exposure, which roughly corresponds to brightness, is controlled by the aperture, ISO speed rating, and exposure time. Increasing the aperture size, ISO speed, or exposure time leads to a brighter image. Another example is the strength of the bokeh effect, which depends on the aperture and focal length.  A larger aperture or longer focal length enhances the bokeh effect. Therefore, accurately identifying numerical camera settings requires a combination of reasoning, implicit knowledge of physics, and visual comprehension of how these settings affect the photographic result, representing a highly complex and intelligent capability. Developing this capability is crucial for practical applications, such as enabling MLLMs as photography assistant agents in real-world scenarios.

Photography-related tasks are challenging for vision language models~(VLMs), such as CLIP~\cite{radford2021clip} and OpenCLIP~\cite{ilharco2021openclip, cherti2023reproducible}. VLMs struggle with numerical concepts~\cite{paiss2023teaching, fang2024camera} and perform poorly when classifying the numerical camera settings based on the images, leading to near-random zero-shot performance~\cite{fang2024camera}. Motivated by these findings, we investigate the following questions: Do MLLMs exhibit visual reasoning capabilities in photography? Specifically, can MLLMs distinguish visual differences across numerical camera settings? 

Inspired by Fang~\etal~\cite{fang2024camera}, we prompt MLLMs~\cite{ku2023viescore,hu2023tifa} to select the most probable camera setting from the given options. Specifically, we evaluate the performance of MLLMs in two scenarios with binary-choice questions: (1) distinguishing between two options of camera settings in which the values of their camera parameters are all different, and (2) distinguishing between two options that differ in the values of only one camera parameter. 
Additionally, we evaluate MLLMs on a more challenging five-choice task, which includes options of camera settings differing in either only one or all camera parameters. 

Our preliminary results, visualized in Figure~\ref{fig:radar-chart}, indicate that while some MLLMs perform well on easier tasks (distinguishing between two completely different camera settings), they still struggle with more challenging visual reasoning tasks, such as identifying the difference of a single camera setting. 
Moreover, the reasoning-focused MLLM outperforms its standard version~(Gemini 2.0 Flash), highlighting the importance of reasoning capabilities in photography-related tasks.
Moreover, no MLLM consistently outperforms others in visual reasoning tasks of photography, demonstrating that photography-related benchmarks are still challenging evaluation tools to assess the visual reasoning capability of MLLMs. 

\begin{table*}[t]
    \centering
    \vspace{-1em}
    \resizebox{0.97\textwidth}{!}{
    \begin{tabular}{c|ccccc|c}
     & \multicolumn{5}{|c|}{Binary Choice} & Five Choice\\
     
    MLLMs & All & Focal Length & Aperture & ISO Speed Ratings & Exposure Time &  All\\
    \hline
    Llama 3.2 11B Vision Instruct & 0.520 & 0.640 & 0.620 & 0.520 & 0.500 & 0.548 \\
    Llama 3.2 90B Vision Instruct (bf16) & 0.680 & 0.710 & 0.630 & 0.550 & 0.460 & 0.589 \\
    Llama 3.2 90B Vision Instruct (fp16) & 0.710 & \textbf{0.720} & 0.660 & 0.520 & 0.440 & 0.547 \\
    Gemini 2.0 Flash & 0.657 & 0.630 & 0.670 & 0.520 & 0.480 & \textbf{0.695} \\
    GPT-4o & 0.780 & 0.580 & 0.490 & 0.480 & 0.510 & 0.444 \\
    GPT-4.5-preview & \textbf{0.859} & 0.580 & 0.490 & 0.480 & 0.510 & 0.551 \\   
    \hline
    Gemini 2.0 Flash Thinking Experimental & 0.727 & 0.639 & \textbf{0.765} & \textbf{0.571} & \textbf{0.687} & 0.579 \\
    OpenAI o1 & 0.850 & 0.570 & 0.500 & 0.470 & 0.520 & 0.646 \\ 
    \hline
    Random Guess & 0.500 & 0.500 & 0.500 & 0.500 & 0.500 & 0.200 \\ 
    \end{tabular}
    }
    \vspace{-0.5em}
    \caption{\textbf{Preliminary results of CameraBench.}
    Binary Choice
    All: The camera settings in the two options are all different.
    Focal Length: Only the focal length is different between the two options.
    Aperture: Only the aperture is different between the two options.
    ISO: Only the ISO speed rating is different between the two options.
    Exposure Time: Only the exposure time is different between the two options.
    Five Choice All: Five-choice tasks with all options.
    }
    \label{tab:main_table}
    \vspace{-0.5em}
\end{table*}

\begin{table*}[ht!]
  \centering
  \scriptsize
  \renewcommand{\arraystretch}{1.5}
  \resizebox{0.98\textwidth}{!}{
  \begin{tabular}{|p{1\linewidth}|}
    \hline
To determine the most likely camera settings used in this image, we need to analyze the characteristics of the scene and how they relate to different camera settings. \\

**Scene Characteristics:** \\

*   The image features a landscape with mountains, trees, and a lake. \\
*   The sky is cloudy, indicating overcast conditions. \\
*   The lighting appears soft and diffused, suggesting that the sun is not directly overhead. \\

**Camera Setting Considerations:** \\

*   **Focal Length:** A wide-angle lens (around 24-35mm) would be suitable for capturing the expansive landscape, including the mountains and the lake. \\
*   **Aperture:** A smaller aperture (higher f-stop number) is often used in landscape photography to ensure a large depth of field, keeping both the foreground and background in focus. \\
*   **ISO:** A lower ISO (around 100-400) is typically preferred for landscape photography to minimize noise and digital artifacts. \\
*   **Exposure Time:** A slower shutter speed can be used to capture the movement of clouds or water, but it might not be necessary in this case given the cloudy sky. \\

**Analysis of Options:** \\

*   **Option A:** Focal length of 27.54mm, aperture of f/14.58, ISO of 524.88, and exposure time of 0.00625 second. \\
\hspace{1cm}    *   The focal length is slightly wider than what might be ideal for a landscape, but it could work depending on the composition. \\ 
\hspace{1cm}    *   The aperture is very small, which would indeed provide a large depth of field but might result in underexposure due to the low light conditions. \\
\hspace{1cm}    *   The ISO is quite high, which could introduce noise in the image. \\ 
\hspace{1cm}    *   The exposure time is very short, which might not capture the desired effect of the clouds or water. \\
*   **Option B:** Focal length of 50.00mm, aperture of f/4.50, ISO of 200.00, and exposure time of 0.0125 second. \\
\hspace{1cm}    *   The focal length is more standard for landscape photography, offering a good balance between wide-angle and telephoto views. \\
\hspace{1cm}    *   The aperture is smaller than what is typically used for landscapes, which could result in underexposure. \\
\hspace{1cm}    *   The ISO is lower, which is beneficial for reducing noise. \\
\hspace{1cm}    *   The exposure time is slightly longer than in Option A, which could help capture more detail in the scene. \\

**Conclusion:** \\

Given the considerations above, **Option B** seems more appropriate for capturing a landscape under overcast conditions. It offers a balance between focal length, aperture, ISO, and exposure time that aligns well with typical landscape photography practices. However, without knowing the exact lighting conditions and the photographer's intent, it's challenging to definitively choose between the two options based solely on the information provided.
    \\ \hline

  \end{tabular}
  }
  \vspace{-0.5em}
  \caption{A failure case of Llama 3.2 11B Vision Instruct.}
  \vspace{-2.5em}
  \label{tab:llama32-11b}
\end{table*}

\begin{table*}[ht!]
  \centering
  \scriptsize
  \renewcommand{\arraystretch}{1.5} 
  \resizebox{0.98\textwidth}{!}{
  \begin{tabular}{|p{1\linewidth}|}
    \hline
**Answer:** (A)\\

**Explanation:** \\

Let's analyze each camera setting option in the context of the provided landscape image: \\

**Option A: Focal length: 27.54mm, Aperture: f/14.58, ISO: 524.88, Exposure Time: 0.00625 second** \\

*   **Focal Length (27.54mm):** This is a wide-angle focal length. Wide-angle lenses are commonly used for landscape photography because they capture a broad field of view, encompassing the vastness of the scene, like the mountains, lake, and trees in the image. This is very suitable for this type of shot.\\
*   **Aperture (f/14.58):** This is a narrow aperture (high f-number). Narrow apertures are used to maximize the depth of field, ensuring that both the foreground and background are in focus. In landscape photography, you typically want everything from the near foreground to the distant mountains to be sharp. f/14.58 is a good choice for achieving a large depth of field.\\
*   **ISO (524.88):** This is a moderate ISO. In slightly overcast or cloudy conditions (as suggested by the image's lighting), a slightly higher ISO than the base ISO (usually 100) might be necessary to maintain a reasonable shutter speed, especially when using a narrow aperture like f/14.58. ISO 524.88 is a reasonable value for such conditions, balancing image brightness with noise levels.\\
*   **Exposure Time (0.00625 second - which is 1/160 of a second):** This is a relatively fast shutter speed. In daylight conditions, even under clouds, a fast shutter speed like 1/160 second is plausible, especially when combined with the aperture and ISO values. It's fast enough to avoid motion blur from hand-holding the camera and to freeze any slight movement in the scene. \\

**Option B: Focal length: 50.00mm, Aperture: f/4.50, ISO: 200.00, Exposure Time: 0.0125 second** \\

*   **Focal Length (50.00mm):** This is a standard focal length, often considered close to human perspective. While it can be used for landscapes, it's less ideal for capturing the expansive view seen in the image compared to a wide-angle lens. A 50mm lens would result in a narrower field of view, potentially cropping out some of the scene's breadth. \\
*   **Aperture (f/4.50):** This is a wider aperture (lower f-number). Wider apertures create a shallower depth of field, meaning only a portion of the image will be in sharp focus, while the foreground and background might be blurred. While shallow depth of field can be artistic, it's generally not desired in landscape photography where you want everything to be sharp. f/4.50 would likely result in less of the scene being in focus, especially with a landscape scene that has elements at varying distances. \\
*   **ISO (200.00):** This is a low ISO, which is good for image quality (less noise). However, in conjunction with a wider aperture like f/4.50 and a relatively fast shutter speed, it might suggest brighter lighting conditions than what is depicted in the image, or it might be compensating for a slower shutter speed (though 0.0125s is still quite fast). \\
*   **Exposure Time (0.0125 second - which is 1/80 of a second):** This is also a fast shutter speed, but slightly slower than Option A. It's still reasonable for daylight conditions. \\

**Why Option A is more likely:** \\

Option A's settings are more aligned with typical landscape photography techniques: \\

*   **Wide focal length** to capture the expansive scene.\\
*   **Narrow aperture** to maximize depth of field and ensure sharpness throughout the image.\\
*   **Moderate ISO** to balance image brightness and noise in slightly less than perfect lighting conditions while using a narrow aperture.\\
*   **Fast enough shutter speed** to avoid motion blur.\\

Option B's settings are less typical for this type of landscape:\\

*   **Standard focal length** is less ideal for capturing the broad landscape. \\
*   **Wider aperture** is counter to the goal of maximizing depth of field in landscapes.\\

Therefore, **Option (A)** is the most likely set of camera settings used to capture this image. \\
  \hline
  \end{tabular}
}
  \vspace{-0.5em}
  \caption{A successful case of Gemini 2.0 Flash Thinking Experimental.}
  \vspace{-2.5em}
  \label{tab:gemini-thinking}
\end{table*}

\section{Data Curation and Benchmark}

Building upon CameraSettings20K~\cite{fang2024camera}, we introduce CameraSettings25K by adding data from MIT-AdobeFiveK~\cite{fivek}, resulting in a total of 25,127 images. As in CameraSettings20K~\cite{fang2024camera}, we normalize camera settings to the standard 35mm (full-frame) format using crop factors. For further details on this normalization process, please refer to the paper of CameraSettings20K~\cite{fang2024camera}.
To benchmark MLLMs, we design six tasks:
(1) binary-choice tasks with two completely different camera settings, 
(2) binary-choice tasks differing only in focal length, 
(3) binary-choice tasks differing only in aperture, 
(4) binary-choice tasks differing only in ISO speed rating, 
(5) binary-choice tasks differing only in exposure time, and 
(6) Five-choice tasks incorporating all options. 
Figure~\ref{fig:sample} shows an example of Task (1) and (6).
The correct answer is obtained from the image's Exif metadata and normalized to the standard 35mm (full-frame) format.
The incorrect answer is randomly sampled from the Exif metadata of a different image in the CameraSettings25K dataset, also
normalized to the standard 35mm format.

Due to resource limitations, we randomly sample 100 images from CameraSettings25K along with their numerical camera settings, including focal length, aperture (f-number), ISO speed ratings, and exposure time. Additionally, we randomly sample another 100 different camera settings from the dataset as incorrect answers. To mitigate selection bias~\cite{pezeshkpour2023large, zheng2023large}, we randomly shuffle the answer choices. To reduce randomness in benchmarking, we set the temperature and Top-P to 0.1 whenever possible. We run inference for Llama 3.2~\cite{meta2024llama32} on our local server, while other MLLMs are accessed via API calls.  Due to computational constraints, we run Llama 3.2 90B Vision Instruct in half-precision floating-point format
(bf16 and fp16), whereas Llama 3.2 11B Vision Instruct runs in single precision (fp32). Despite our best efforts, some MLLM outputs cannot be identified. We remove these samples, resulting in some MLLMs being evaluated with fewer than 100 samples.
\section{Experimental Results}

\textbf{Quantitative Results.} The experimental results are presented in Table~\ref{tab:main_table} and visualized as radar charts in Figure~\ref{fig:radar-chart}. We observe that the scaling law~\cite{kaplan2020scaling} also applies to Task (1); performance improves as model size increases. However, in Task (2), (3), (4), and (5), most models perform close to random guessing. 
For Task (6), which involves five-choice questions, models achieve better performance compared with Task (2), (3), (4), and (5). 
Furthermore, the reasoning-focused MLLMs (Gemini 2.0 Flash Thinking Experimental and OpenAI o1) outperforms the regular MLLMs (Gemini 2.0 Flash, GPT-4o, and GPT-4.5) in many tasks, indicating the critical role of visual reasoning capability in photography-related tasks. 
These observations suggest that current models are limited in their ability to handle visual reasoning tasks, indicating the gap for improvement. 

\noindent \textbf{Qualitative Results.} Due to the page limit, we present one illustrative case comparing a regular MLLM (Llama 3.2 11B Vision Instruct) in Table~\ref{tab:llama32-11b} and a reasoning-focused MLLM (Gemini 2.0 Flash Thinking Experimantal) in Table~\ref{tab:gemini-thinking}. The visual and textual inputs for this case are provided in the Task (1)'s prompt of Figure~\ref{fig:sample}. The correct answer is (A).
We specifically compare the results with Llama 3.2 11B Vision Instruct rather than Gemini 2.0 Flash because Gemini 2.0 Flash directly outputs the incorrect final answer.
We exclude GPT-4o and OpenAI o1 from this comparison because the OpenAI API does not provide intermediate reasoning tokens, and directly outputs the final answer. 

Overall, the reasoning-focused MLLM provides more detailed explanations. In contrast, the regular MLLM exhibits hallucinations, incorrectly stating that a 50mm focal length is standard for landscape photography. Additionally, the regular MLLM provides irrelevant information, such as stating, ``The aperture is smaller than what is typically used for landscapes,'' even though the aperture in the option is actually larger than the other option. This comparison shows the effectiveness of reasoning-focused MLLMs for this visual reasoning task, implicitly highlighting that identifying numerical camera settings requires reasoning capabilities.
\section{Conclusions}
In this ongoing work, we introduce CameraBench, a novel benchmark for visual reasoning. We use photography-related tasks, identifying numerical camera settings, to evaluate the visual reasoning capabilities of multimodal large language models~(MLLMs). Successfully addressing these tasks requires reasoning, implicit understanding of physics, and interpreting visual differences tied to numerical concepts. Our preliminary results, based on 100 samples from CameraSettings25K, demonstrate the effectiveness of this benchmark for evaluating MLLMs' visual reasoning capabilities. Future work will involve more extensive benchmarking and a comprehensive evaluation using the entire CameraSettings25K dataset across various MLLMs.

\section*{Acknowledgement}
This research is supported by National Science and Technology Council, Taiwan (R.O.C), under the grant number of NSTC-113-2634-F-002-007, NSTC-112-2222-E-001-001-MY2, NSTC-113-2634-F-001-002-MBK, 113-2634-F-002-008- and Academia Sinica under the grant number of AS-CDA-110-M09. 

{
    \small
    \bibliographystyle{ieeenat_fullname}
    \bibliography{main}
}

\end{document}